\documentclass{article}

\usepackage{microtype}
\usepackage{graphicx}
\usepackage{subfigure}
\usepackage{booktabs} % for professional tables

\usepackage{hyperref}

\usepackage[accepted]{icml2020}

\usepackage{amsmath, amssymb, multirow, colortbl}
\usepackage[utf8]{inputenc}
\usepackage[T1]{fontenc}
\icmltitlerunning{Symbolic Regression using MINO}

\begin{document}

\twocolumn[
\icmltitle{Symbolic Regression using Mixed-Integer Nonlinear Optimization}

% It is OKAY to include author information, even for blind
% submissions: the style file will automatically remove it for you
% unless you've provided the [accepted] option to the icml2020
% package.

% List of affiliations: The first argument should be a (short)
% identifier you will use later to specify author affiliations
% Academic affiliations should list Department, University, City, Region, Country
% Industry affiliations should list Company, City, Region, Country

% You can specify symbols, otherwise they are numbered in order.
% Ideally, you should not use this facility. Affiliations will be numbered
% in order of appearance and this is the preferred way.
\icmlsetsymbol{equal}{*}

%\author{,\textsuperscript{\rm 1} 
%,\textsuperscript{\rm 1} 
%,\textsuperscript{\rm 1} 
%,\textsuperscript{\rm 1}
%,\textsuperscript{\rm 2}
%\textsuperscript{\rm 1}\\

\begin{icmlauthorlist}
\icmlauthor{Vernon Austel}{wat}
\icmlauthor{Cristina Cornelio}{wat}
\icmlauthor{Sanjeeb Dash}{wat}
\icmlauthor{Jo\~ao Gon\c calves}{wat}
\icmlauthor{Lior Horesh}{wat}
\icmlauthor{Tyler Josephson}{umin}
\icmlauthor{Nimrod Megiddo}{alm}
\end{icmlauthorlist}

\icmlaffiliation{wat}{IBM Research AI, 1101 Kitchawan Road, Yorktown Heights, NY 10598}
\icmlaffiliation{umin}{Department of Chemistry, University of Minnesota}
\icmlaffiliation{alm}{IBM Research - Almaden, 650 Harry Rd, San Jose, CA 95120}

\icmlcorrespondingauthor{Sanjeeb Dash}{sanjeebd@us.ibm.com}

% You may provide any keywords that you
% find helpful for describing your paper; these are used to populate
% the "keywords" metadata in the PDF but will not be shown in the document
\icmlkeywords{Machine Learning, ICML}

\vskip 0.3in
]

% this must go after the closing bracket ] following \twocolumn[ ...

% This command actually creates the footnote in the first column
% listing the affiliations and the copyright notice.
% The command takes one argument, which is text to display at the start of the footnote.
% The \icmlEqualContribution command is standard text for equal contribution.
% Remove it (just {}) if you do not need this facility.

\printAffiliationsAndNotice{}  % leave blank if no need to mention equal contribution
%\printAffiliationsAndNotice{\icmlEqualContribution} % otherwise use the standard text.

\begin{abstract}
The Symbolic Regression (SR) problem, where the goal is to find a regression function that does not have a pre-specified form, but is any function that can be composed of a list of operators, is a hard problem in machine learning, both theoretically and computationally. Genetic programming based methods, that heuristically search over a very large space of functions, are the most commonly used methods to tackle SR problems. An alternative mathematical programming approach, proposed in the last decade, is to express the optimal symbolic expression as the solution of a system of nonlinear equations over continuous and discrete variables that minimizes a certain objective, and to solve this system via a global solver for mixed-integer nonlinear programming problems. Algorithms based on the latter approach are often very slow. We propose a hybrid algorithm that combines mixed-integer nonlinear optimization with explicit enumeration and incorporates constraints from dimensional analysis. We show that our algorithm is competitive, for some synthetic data sets, with a state-of-the-art SR software and a recent physics-inspired method called AI Feynman.
\end{abstract}

\section{Introduction}

%Discovering mathematical models of physical, biological, or economic systems is a difficult intellectual endeavor. Model discovery  based on first principles, where data is primarily used for model verification and refinement, can sometimes  lead to models with remarkable levels of universality and interpretability. This is a human-centric process, and not scalable. Automated, data-driven approaches to model discovery have now become ubiquitous, as they are more scalable, and can be rapidly deployed. However, they often need a lot of data, and yield non-interpretable models that cannot be transferred to a broader setting. Generalizability and interpretability require a priori imposition of a statistically and computationally parametric form on the model, and discovery amounts to parameter estimation or regression.

Regression methods have numerous applications in science and economics. In traditional regression methods, such as linear or logistic regression, the relationship between a dependent variable -- say $y$ -- and a number of independent variables -- say $x_1, \ldots, x_n$ -- is assumed to have a fixed, parametric, functional form, and the parameters are calculated from data to minimize a specific ``loss'' function. For example, in linear regression, the functional form is assumed to be a linear function of the independent variables: $y = a_1x_1 + \cdots + a_nx_n$. The unknown parameters $a_i$ are the coefficients that are calculated so as to minimize, say, the least-square error: $\sum_i (y^i - \sum_j a_jx_j^i)^2$, where $y^i$ and $x_j^i$ are the values of $y$ and $x_j$, respectively, in the $i$th data point.

In symbolic regression (SR), both the functional form of of a regression function, and the associated parameters or coefficients are calculated from data so as to minimize some loss function. The functional form is assumed to be anything that can be composed from a given list of basic functions or operators applied to the independent variables and arbitrary constants. For example, if the operators are  $+$, $-$ and $\times$, then the space of all possible functions is the set of all polynomials of arbitrary degree.%, and therefore polynomial regression can be viewed as a special case of symbolic regression.
The usefulness of SR was demonstrated in
\cite{connor1977scaling,langley1981data,willis1997genetic,davidson2003symbolic,schmidt2009distil,schmidt2010symbolic}.

Given its generality, symbolic regression is clearly a difficult problem and there has been a lot of research into designing computationally effective algorithms.
%Heuristics to find explicit functional relationships were developed in the BACON system \cite{langley1987}.
Starting with \cite{koza}, SR problems were solved with genetic programming (GP) \cite{augusto2000symbolic,banzhaf1998genetic}, and SR is often treated as being synonymous with GP \cite{gplearn}.
In \cite{korns}, the accuracy of SR solutions achieved by genetic algorithms was claimed to be poor. "Bloat" is another common issue in GP based methods (obtaining functions with high description length).

Some recent research on improving symbolic regression techniques is aimed at 
speeding them up \cite{luo2017}, 
finding accurate constants in the derived symbolic mathematical expression \cite{dgp},
and searching for implicit functional relationships \cite{schmidt2010symbolic}. 
These approaches are primarily based on genetic algorithms. 
Another approach explicitly populates a large hypothesis space of functions, on which sparse selection is applied \cite{brunton2016discovering}. % Since the candidate expressions are non-linear, sparse relaxation embedding theory does not apply. While these heuristics have undoubtedly been successful and useful, they possess limited mathematical rigor, and are limited in their potential reach and analyzability. Lastly, a recent approach relies on reservoir neural network training which potentially handle chaotic behaviors, {\bf add citation [7]}.
In \cite{AI-Feynman}, a non-GP SR algorithm is developed , potentially well-suited for physics problems, that combines neural-network fitting with several techniques inspired by physics, for example, dimensional analysis, and also complete enumeration of symbolic expressions of a certain limited size.

A major challenge in symbolic regression is the difficulty of finding scientifically meaningful models out of the large number of possible models that may fit given data.
In \cite{schmidt2009distil,schmidt2010symbolic}, partial derivatives are matched, in addition to fitting data, to discover meaningful functions. In AI Feynman \cite{AI-Feynman}, dimensional analysis is used to both speed up the solution process, and to generate meaningful physical models.

Some recent papers proposed mathematical optimization based approaches for obtaining regression models: a model is obtained as a solution of a constrained optimization problem that selects the kind of model and optimizes its parameters. 
Various Mixed-Integer Programming (MIP) formulations were proposed in \cite{bertsimas} for finding classification and regression models.
Mixed-Integer Nonlinear Programming (MINLP) formulations for SR were introduced in 
\cite{cozadthesis}, 
\cite{horesh2015globally} and
\cite{horesh2016tech}. Some computational results were presented in \cite{cozadthesis,Austel2017,cozsah}. In these three papers,
the BARON \cite{ts05} solver was used to obtain globally optimal (see the next section) solutions, though other similar solvers can be used.
An advantage of the MINLP framework is that, in principle, globally optimal symbolic expressions can be obtained along with certificates of optimality. However, existing MINLP models for SR, and the solvers for such problems, have limited scalability. 
In addition, the MINLP framework allows one to incorporate domain knowledge that can be encoded via nonlinear constraints: adding constraints to enforce dimensional consistency is proposed in  \cite{horesh-dim}.

\subsection{Our contribution}

In this paper, we develop a symbolic regression algorithm based on solving MINLPs inspired by and extending the free-form SR discovery framework \cite{Austel2017} that was tested on a few toy problems, such as relations found by Kepler and Galileo.
Our work is different from the work in the previous paragraph in multiple ways. Firstly, we sacrifice global optimality to get a faster algorithm (thus our algorithm can be viewed as a heuristic, though we can, in principle obtain globally optimal solutions given enough time).
Secondly, we choose a different MINLP formulation of an expression tree than in previous papers, and also employ a different search strategy. Finally, we add constraints to enforce dimensional consistency.

We compare our method with  AI Feynman \cite{AI-Feynman} on the synthetic, physics-inspired data set in the associated paper, and show that our MINLP-based approach is still competitive for a number of problems, and obtains solutions to some problems that are not solved by Eureqa (but solved by AI Feynman). We can also obtain solutions to some problems from less data. %Finally, we also study problems with real, experimental data, where the desired functional forms have complicated, unknown constants, and are thus beyond the reach of AI-Feynman (and many SR packages).

%%%%%%%%%%%%%%%%%%%%%%%%%%%%%%%%%%%%%%%%%%%%%%%%%%%%%%%%%%%%

\section{Related work}
%%%%%%%%%%%%%%%%%%%%%%%%%%%%%%%%%%%%%%%%%%%%%%%%%%%%%

\begin{figure}
\begin{center}
\includegraphics[width=\columnwidth]{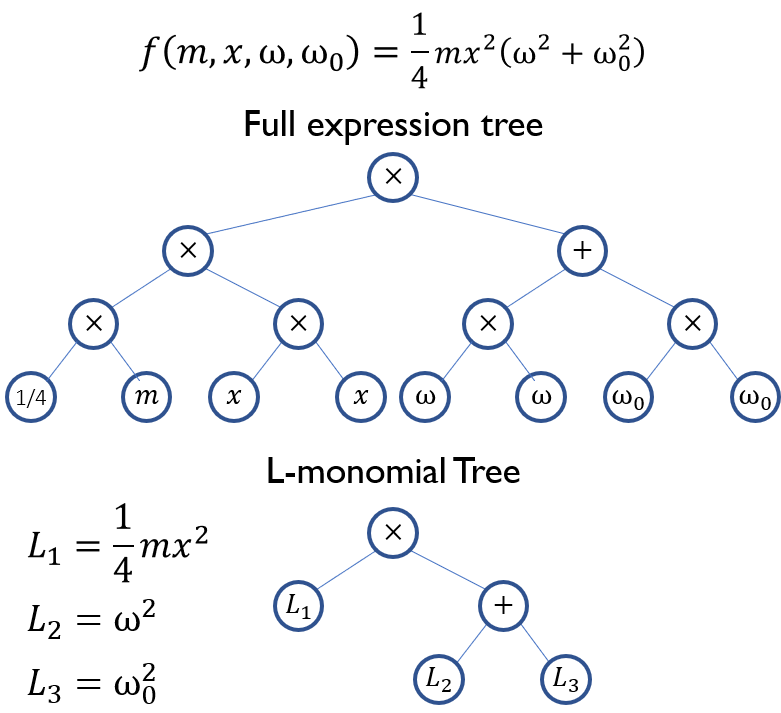}
\caption{Expression tree for $\frac{1}{4}mx^2(\omega^2 + \omega^2_0)$, expressed with full arithmetic notation, and as a more compact tree of L-monomials.}
\label{fig:expression_tree}
\end{center}
\end{figure}

A symbolic regression scheme consists of a space of valid mathematical expressions composable from a basic list of operators (we assume these to be unary or binary), and a mechanism for exploring the space.  
Each valid mathematical expression can be represented by an expression tree: a rooted binary tree where each non-leaf node has an associated binary or unary operator ($+$, $-$, $\times$, $\sqrt{}$, $\log$, etc.), and each leaf nodes has an associated constant or independent variable. 
An example of an expression tree for the expression
%$$ e^{\sqrt{3x}} + \frac{1}{2}\sqrt{\frac{y}{z}} $$
$$ \frac{1}{4}mx^2(\omega^2 + \omega_0^2)$$
is presented in Figure~\ref{fig:expression_tree}. The second tree in the figure is explained later.

%\subsection{Genetic Programming}

A typical genetic programming (GP) solver searches the space of expressions using a genetic algorithm
%by maintaining a population of expressions, starting from a set of randomly selected ones. It creates random ``mutations'' (modified subexpressions) and recombines subexpressions, while a set of the fittest expressions seen so far. The ``fitness'' of an expression is a function of its complexity and predictive accuracy.
The Eureqa package \cite{schmidt14}, based on the article \cite{schmidt2009distil}, is a state-of-the-art GP-based solver. Another popular solver is {\tt gplearn} \cite{gplearn}.
%There have been many recent papers published on GP algorithms for symbolic regression, and we cannot mention all of them here \tyler{citing a review would make sense here}.
AI Feynman \cite{AI-Feynman} %is geared towards problems that have high-quality data, and where the dimensions of all quantities are specified.
first trains a neural network (NN) on the input data, so that the function encoded by the NN is an accurate fit for the input data. %In other words, they create a potentially complicated encoding of the function to be discovered.
It then uses tests if the true function has various properties such as separability, symmetry etc, by evaluating the NN at points not originally in the data set. In addition, it uses dimensional analysis to speed up the search for the best symbolic expression. 
In \cite{AI-Feynman}, the authors applied their algorithm to one-hundred physics equations, and claim to rediscover all of them. They observe that
Eureqa obtained solutions to only 71 of these instances.

\newcommand{\Z}{\mathbb{Z}}
\newcommand{\R}{\mathbb{R}}
\subsection{Globally optimal symbolic regression}
In a series of works \cite{cozadthesis,cozsah,horesh2015globally,horesh2016tech,Austel2017}, 
symbolic regression (SR) methods based on combined discrete and continuous optimization were developed, and the symbolic regression problem was formulated in various ways as a Mixed-Integer Nonlinear Programming (MINLP) problem. The MINLP problem is solved to global optimality using an off-the-shelf MINLP solver such as BARON \cite{ts05}, COUENNE  or SCIP \cite{MaherFischerGallyetal2017}.
These solvers solve problems of the form
\begin{eqnarray}
\min & f(x,y) \\
s.t. & g(x,y) \leq b \\
& x \in \R^m, y \in \Z^m
\end{eqnarray}
where $x$ is a vector of $m$ continuous variables, $y$ is a vector of $n$ discrete variables, $f(x,y)$ is a real-valued function that can be composed using a finite list of operators such as $+, -, \times, /, exp(\cdot)$ etc. (this list will vary with each solver), and $g(x,y)$ is a vector-valued function created using the same operators.
These solvers
use various convex-relaxation schemes to obtain lower bounds on the objective-function value, and utilize these bounds in branch-and-bound schemes to obtain globally optimal solutions.  
%\marginpar{If these numbers refer to the MINLP index, what is 21?}

%In \cite{cozsah}, the objective function is the sum of squared deviations (of the values of the derived symbolic expression from the output values in data), and the model complexity is controlled simply by a bound on the number of active nodes in the expression tree. In other formulations, for example \cite{Austel2017}, more refined measures of model complexity are incorporated into the objective function, while imposing an upper bound on accuracy, measured as the sum of squared deviations. 

This approach produces a globally-optimal mathematical expression (along with a certificate of optimality) while avoiding an exhaustive search of the solution space. Another advantage is that it directly produces correct, within a tolerance, real-valued constants; most other methods use specialized algorithms to refine constants \cite{dgp} and cannot guarantee global optimality. 
%Moreover, the optimization formulation may allow straightforward incorporation of domain knowledge, via constraints associated with conservation, symmetry, dimensional consistency, etc., or user preferences.
%We note that in \cite{badek} the authors use constraints to restrict the search space but this approach is different from ours as the search there is by a genetic algorithm.

%We note that the space of all possible mathematical models, i.e., expression trees and their parameters, is very large, and a brute-force search is obviously not a feasible approach. However, Branch-and-Bound provides a framework within which a number of clever techniques can be applied quite successfully. For example, instances of the famously NP-hard Traveling Salesman Problem with $10^5$ cities can now be provably solved to global optimality, despite the combinatorial explosion that leads to more than $10^5!$ possible solutions. 

In these MINLPs, the set of valid binary expression trees is specified by a set of constraints over discrete and continuous variables. The discrete variables of the formulation are used to define the structure of the expression tree including the assignment of operators to non-leaf nodes and whether a leaf node is assigned a constant or a specific independent variable. The continuous variables are used for the undetermined constants, and also to evaluate the resulting symbolic expression for specific numerical values associated with individual data points.
The objective functions vary from accuracy (measured as sum of squared deviations) to model complexity in \cite{Austel2017}.

In \cite{Austel2017}, for example, the MINLP has the following form:
\begin{eqnarray*}
\min~~~~~~ &  \mathcal{C}(f_{\theta}) & \mbox{complexity}\\ 
\mbox{s.t.}~~~~~~~ & \theta \in \mathcal{T} &  \mbox{grammar}\\ 
& v_i = f_{\theta}(x^{(i)}), ~~ \forall i \in I  & \mbox{prediction}\\
& \mathcal{D}(f_{\theta}(x),y) \leq \varepsilon & \mbox{error}
%& f_{\theta} \in V & \mbox{invariance}
\end{eqnarray*}
where $f_{\theta}$ represents an expression tree defined by the structural and continuous decision variables collectively designated as $\theta$; $\mathcal{T}$ is the universe of valid expression trees; $\mathcal{C}$ measures the description complexity; $\mathcal{D}$ measures error of the predicted values $v$; $y$ is the vector of observations for inputs $\{x^{(i)}\}_{i \in I}$.
%and $V$ a set of invariances that qualifies the expression as viable.

\section{System Description}\label{sec-desc}
%The technique we explored to reduce the search space is dimensionality analysis: the idea is especially useful for discovering equations in physics, where each input has an associated physical dimension. The dimensions help reduce the search space considerably because only a small number of possible combinations of inputs are consistent with the dimensions. 
We first describe the basics of our system (without dimensional analysis), and then later explain how we incorporate dimensional analysis.

We take as input a list of operators (for this discussion, assume the input operators are $+, -, \times,/$, and $\sqrt{}$), an upper bound $d$ on the tree depth, an upper bound $k$ on the number of constants, and a domain for the constants $[-\Omega , \Omega]$. In Figure~\ref{fig:expression_tree}, there is a single constant (distinct from 1) with value $1/4$ in the expression tree and in the L-monomial tree.
In the prior work on globally optimal symbolic regression, the MINLP formulations typically have discrete variables that (1) determine the placement of the operators in nodes of the expression tree and (2) the mapping of independent variables to leaf nodes, and (3) determine whether to map an independent variable or a constant to a leaf node.

In our formulation, we simply do not have discrete variables for (1). We explicitly enumerate all possible assignments of the operators in expression trees up to depth $d$. More precisely, if a``partial'' expression tree is one where the leaf nodes are undetermined, but the operator assignments to non-leaf nodes are determined, then we enumerate all possible partial expression trees up to a certain depth.
Our assignment of variables/constants to leaf nodes is also different from prior work. Let $x_1, \ldots, x_n$ be all the independent variables. Instead of assuming that each leaf node is either a constant $h$ or one of $x_1, \ldots, x_n$ as in the prior work, we assume each leaf node is a one-term multivariable Laurent polynomial of the form
\begin{equation}\label{lmonomial} hx_1^{a_1}x_2^{a_2}\cdots x_n^{a_n}, \end{equation}
where $a_1, \ldots, a_n$ are (undetermined) integers (for computational efficiency, we limit these integers to lie in the range $[-\delta, \delta]$ for some input constant $\delta$), and $h$ represents an (undetermined) constant in the final expression.
We refer to an expression of the type (\ref{lmonomial}) an {\it L-monomial}.
In other words, rather than assigning a single variable or constant to a leaf node, we potentially assign both variables and constants, and also multiple variables (with positive or negative powers to a leaf node).  We call the resulting trees {\em generalized expression trees}, and {\em gentrees} for convenience.
%a rooted binary tree with depth of at most $d$ (as measured from the root), where each non-leaf node has an associated operator from the list of input operators, and each leaf node corresponds to an undetermined L-monomial in terms of the independent variables.

Each gentree $T$ (with depth say $d$) corresponds to a symbolic expression: each non-leaf node with height $1$ corresponds to the expression formed by applying the operator at the node to the expressions (i.e., L-monomials) in the children nodes, and non-leaf nodes at greater heights are handled in the same manner in order of height. The only gentree with depth 0 corresponds to the L-monomial $L_1$, whereas the gentrees of depth 1 correspond to the expressions $\sqrt{L}$, $L_1+L_2$,\, $L_1 \times L_2$,\, $L_1/L_2$ and $L_1-L_2$, respectively, where $L_1$ and $L_2$ are L-monomials.

\subsection{Pruning the list of gentrees}

We try to reduce the number of relevant partial expression trees/gentrees by removing a number of "redundant" trees using the fact that
the set of nonzero (the constant $h$ in \ref{lmonomial} is nonzero) L-monomials is closed under multiplication and division. 
Notice that both $L_1\times L_2$ and $L_1/L_2$ are L-monomials and thus can be represented by a single expression $L_1$. In other words, if there is a symbolic expression $f$ of the form $f = L_1\times L_2$ that fits our data, then there is one of the form $f = L_1$.
Accordingly our first few pruning rules are: remove a tree $T$ from the list $\mathcal{T}$ of all gentrees with depth up to $d$ if $T$ has the subexpression
\begin{eqnarray*}
    & \mbox{[R1]}  ~~~& L_1 \times L_2 \mbox{ or } L_1/L_2.\\ 
    & \mbox{[R2a]} ~~~& L_1\times (L_3 \pm L_4).\\
    & \mbox{[R2b]} ~~~& (L_1 \pm L_2)*(L_3 \pm L_4). \\
    & \mbox{[R3]}  ~~~& (L_1 \pm L_2)/L_3.
\end{eqnarray*}
The first rule was explained above. The second follows by associativity: $L_1*(L_3 + L4) = L_1\times L_3 + L1\times L_4 = L_1' + L_2'$, for some L-monomials $L_1'$ and $L_2'$. If a $T$ with a subexpression $L_1*(L_3 + L4)$ appears in $\mathcal{T}$, then replacing this subexpression by $L_1' + L_2'$ results in another gentree $T'$ of the same depth, which must therefore be in $\mathcal{T}'$.
The same idea can be applied to the subexpression $L_1*(L_3 - L_4)$.
We can apply associativity twice to justify rule R2b, as $(L_1 + L_2)*(L_3+ L_4) = L_1' + L_2' + L_3' + L_4'$, for some $L_i'$ (the same argument holds when either $+$ is replied by a $-$ in R2b).
Once again, the second expression has the same depth as the first, and therefore there must be a tree in $\mathcal{T}$ containing it. R3 can be explained similarly.

\subsection{MINLP formulation for a gentree}

Let the data points be $x^{(i)}$ for $i \in I$, where $I$ is an index set, and let the features/independent variables be $x_1, \ldots, x_n$.
%Let the value of feature $j$ in the $i$th data point be $x^{(i)}_j$.
Let $y$ stand for the observations/dependent variable, with $y^{(i)}$ standing for the $i$th observation (of the dependent variable).
Let $T$ be a given gentree with $m$ leaf nodes, and let $p$ be the vector of all integer variables corresponding to the powers of the independent variables in the different leaf nodes. Then $p \in \mathbb{Z}^{mn}$. Let $h_1, \ldots h_m$ be the variables corresponding to the constants in leaf nodes $1, \ldots, m$.
Finally, assume we have a 0-1 variable $z_i$ that determines whether or not the $i$th leaf node has a constant $h_i$ that is different from one.
Thus the (vector) variables in our model are $p, z$ and $h$. Let $f_{h,p, z, T}$ stand for the symbolic expression defined by fixing the values of these variables.
Then the MINLP we solve can be framed as
\begin{eqnarray}
\min~~& \sum_{i \in I}(y^{(i)} - f_{h,p,z,T}(x^{(i}))^2  \label{minlp1}\\ 
\mbox{s.t.}~~&  -\delta \leq p_i \le \delta ~~~ \mbox{for } i=1, \ldots, mn \label{power1}\\ 
& -\Omega z_i + (1-z_i) \leq h_i \leq \Omega z_i + (1-z_i) \nonumber\\
& ~~~~~~~\mbox{for } i =1,\ldots, m   \label{const1} \\
& \sum_{i=1}^m z_i \leq k  \label{maxconst1}\\
& z \in \{0,1\}^m, ~~p \in \Z^{mn}  \label{integrality}
\end{eqnarray}

The constraints (\ref{integrality}) and (\ref{power1}) force $z_i$ to take on 0-1 values, and $p_i$ to take on values in the range $\{-\delta, -\delta+1, \ldots, \delta\}$. The constraint (\ref{const1}) restricts $h_i$ to lie in the range $[-\Omega, \Omega]$ if $z_i$ has value 1, and forces $h_i$ to take on value 1, when $z_i$ has value 0.
The constraints (\ref{maxconst1}) allow at most $k$ of the $z_i$ variables to have value 1, and therefore at most $k$ of the $h_i$ values to be different from 1.
The function $f_{h,p,z,T}$ is composed of the operators in the non-leaf nodes of $T$ from the L-monomials in the leaf nodes.
The objective function is the sum of squares of differences between the symbolic expression values for each input data point and the corresponding dependent variable value (call it the {\em least-square error}).

We use BARON to solve the MINLPs we generate, and thus $T$ and $f_{h,p,z,T}$ are limited by the operators that BARON can handle ($+,-,\times,/,\exp(),\log()$).
We will illustrate $f$ for a few examples, rather than specify it formally.
Suppose we are trying to derive the formula $F=G\, \frac{m_1 m_2}{r^2}$, where
$m_1, m_2$ and $r$ are independent variables, and $G$ is an unknown constant, and we have multiple data points (for $i \in I$) with values for the independent variables, and for associated $F$.
If $T$ is a depth 0 gentree, then $T$ is just a single L-monomial,
and $f_{h,p,z,T} = hm_{1}^am_{2}^br_i^c$ where $a,b,c$ are undetermined integers in the range $[-\delta,\delta]$, and $h$ is an undetermined real number in the range $[-\Omega, \Omega]$. The objective function is then
$$\sum_{i\in I} (F_i-hm_{1i}^a m_{2i}^b r_i^c )^2~, $$
where 
$F_i, m_{1i}, m_{2i}$ and $r_i$ are the values of $F, m_1, m_2, r$ in the $i$th data point.
If the gentree can be written as $L_1 + L_2$, then the objective function becomes
$$\sum_{i\in I} (F_i -(h_1m_{1i}^{a_1} m_{2i}^{b_1} r_i^{c_1}+ h_2m_{1i}^{a_2} m_{2i}^{b_2} r_i^{c_2}))^2; $$
here $a_i, b_i,c_i$ and $h_i$ are undetermined, and we solve for these values.

As depicted in Figure~\ref{fig:expression_tree}, an expression tree of a certain depth can sometimes be represented by a gentree of smaller depth. Therefore, by enumerating gentrees of a certain depth, one obtains a much richer class of functions than can be obtained by expression trees of the same depth.

\subsection{Enumeration and parallel processing}

By enumerating the generalized expression trees and solving them separately, the problem is divided into multiple, easier-to-solve sub-problems (operator placement in non-leaf nodes does not have to be determined any more) that can be solved much more quickly. This "divide and conquer" formulation can obtain solutions to problems that were intractable for the formulation in \cite{Austel2017}.

When available, we exploit parallelism, and run multiple threads, with one MINLP corresponding to a single gentree in a single thread.
If we obtain a solution that fits the data within a prescribed tolerance from a gentree with depth $d'$, then we stop all processes/threads containing gentrees with depth $> d'$.  We terminate a gentree if the lower bound on the least-square error exceeds the least-square error found from other gentrees of the same or lower depth. Thus we execute a branch-and-bound type search, and implicitly search for the least depth gentree that fits the data.

If we have more gentrees than available cores, we process them in a round robin fashion; if $t$ is the number of cores, we start solving for $t$ gentrees in parallel, and after a fixed amount of time (10 seconds), we pause the first $t$ gentrees, and start solving $t$ more, till we either find a solution or run out of gentrees in which we case we start from the first gentree. The gentrees are sorted by a measure of complexity (roughly equal to the number of nodes).
Of course, the gentree enumeration (and thus our overall algorithm) grows exponentially with gentree depth $d'$: the number of gentrees for $d'=2, 3, 4, 5$ are, respectively, 7, 60, 4485 and over 100,000. Our gentree enumeration approach is unlikely to be tractable for $d' >= 6$ and is already hard for $d'=4$ if we include more operators than listed earlier.
But each L-monomial can represent large expression trees.

\subsection{Dimensional analysis}
%Two key challenges in symbolic regression are: ({\it i}) the combinatorial explosion of the size of the space of valid expressions with increasing expression length, and ({\it ii}) the difficulty of finding scientifically meaningful models.
Consider Newton’s law of universal gravitation which says that the gravitational force between two bodies is proportional to the product of their masses and inversely proportional to the square of the distance between their centers (of gravity): 
$$F \propto \frac{m_1 m_2}{r^2} $$
and the constant of proportionality is G, the gravitational constant. Therefore, we have
$$F=G\, \frac{m_1 m_2}{r^2}.$$
The units of $G$ are chosen so that units of the right-hand-side expression equal the units of force (mass $\times$ distance $/$  time-squared).
Suppose one is given a data set for this example, where each data item has the masses of two bodies and the distance and gravitational force between them.
Dimensional analysis would rule out $F = m_1m_2/r^2$ or $F = m_1/m_2 + m_2r$, for example, as possible solutions.%, even if these expression fit the data. In the first case, the units of the r.h.s do not match those of $F$ and in the second case, the units of $m_1/m_2$ do not match those of $m_2r$, and the units of either do not match those of $F$.

If constants can have units and every L-monomial can have a such a constant, then dimensional analysis conveys essentially no information.
For example, we can choose constants $h_1, h_2, h_3$ with appropriate dimensions so that both the expressions $F = h_1m_1m_2/r^2$ and $F = h_2m_1/m_2 + h_3m_2r$ satisfy all dimensional requirements.
In the AI Feynman experiments, the authors do not allow constants to have units, and also ensure that for all required universal constants that have units (such as the Gravitational constant), both numerical values and units are given as input.
We allow constants to have units or not based on an input flag.

We next explain the constraints we add to our formulation to enforce dimensional consistency.
Assume that all constants have no units, and that $k=2$, and $\delta = 2$, and $d=1$.
%We demonstrate the usefulness of our approach with numerical experiments. 
For the gravitation example (without the gravitational constant $G$ as an input), each L-monomial $L$ has the form
\begin{equation}L= h m_1^a m_1^b r^c,\label{lmon2}\end{equation}
where $a$, $b$ and $c$ are bounded integer variables with an input range of $[-2,2]$, and $h$ is a variable representing a constant.
We add to the system of constraints (\ref{power1}) - (\ref{integrality})
linear constraints that equate the units of the symbolic expression to those of the dependent variable. For the case our gentree consists of a single node (as in the first tree in Figure 1), our symbolic expression has the form (\ref{lmon2}), and we add the linear constraints
$$
 a 
\begin{bmatrix}
 1 \\ 0\\ 0
\end{bmatrix} 
+ b 
\begin{bmatrix}
 1 \\ 0\\ 0
\end{bmatrix} 
+ c 
\begin{bmatrix}
 0 \\ 1\\ 0
\end{bmatrix} 
=
\begin{bmatrix}
 ~~~1 \\ ~~~1\\ -2
\end{bmatrix} .
 $$
Here, each component of a column vector corresponds to a unit (mass, distance, time, respectively).
The column on the right-hand side represents the units of force, i.e., mass times distance divided by squared time. The columns on the left-hand side represent the dimensionalities of mass and distance, respectively.
The third of the above linear equations does not have a solution. 

Next, assume the symbolic expression is $L_1+L_2$ where $L_1=h_1m_1^{a_1} m_2^{b_1} r^{c_1}$ and $L_2=m_1^{a_2} m_2^{b_2 } r^{c_2}$, and the unkowns are $a_i, b_i, c_i$ and $h_i$. The units of $L_1$ and $L_2$ have to match, and must also equal the units of $F$. The linear constraints we add are
\begin{align}
%\begin{bmatrix}
% ~~~1 \\ ~~~1\\ -2
%\end{bmatrix} 
%&= 
a_1 \label{eq:constraints1}
\begin{bmatrix}
 1 \\ 0\\ 0
\end{bmatrix} 
+ b_1 
\begin{bmatrix}
 1\\ 0\\ 0
\end{bmatrix} 
+ c_1 
\begin{bmatrix}
 0 \\ 1\\ 0
\end{bmatrix} 
&= \begin{bmatrix}
 ~~~1 \\ ~~~1\\ -2
\end{bmatrix} 
\\
%&= 
a_2 \label{eq:constraints2}
\begin{bmatrix}
 1 \\ 0\\ 0
\end{bmatrix} 
+ b_2 
\begin{bmatrix}
 1 \\ 0\\ 0
\end{bmatrix} 
+ c_2 
\begin{bmatrix}
 0 \\ 1\\ 0
\end{bmatrix} 
&=
\begin{bmatrix}
 ~~~1 \\ ~~~1\\ -2
\end{bmatrix}. 
\end{align}
%and the restrictions that $a_1$, $b_1$, $c_1$, $a_2$, $b_2$, $c_2$ are all integral in the range $[-2,2]$.

Finally, for the symbolic expression $L_1 \times L_2$, to compute the units of this expression we need to add up the units of $L_1$ and of $L_2$. Thus we sum the left hand sides of (\ref{eq:constraints1}) and (\ref{eq:constraints2}) and equate this sum to the righ hand side of (\ref{eq:constraints1}).
We can similarly deal with dimension matching in the remaining gentrees of depth 1 via linear constraints.
For greater depth gentrees, we apply the ideas above to depth 1 (non-leaf nodes), and then to depth 2 nodes and so on.

%, the number of decision variables is rather small (e.g., it is equal to $6$ in the above example) and the range is small (equals $5$ in the above example).
% Expressions with exponentiation or logarithmic operations may also leverage dimensional analysis, requiring the arguments of these operations to be dimensionless, but we do not experiment with these operators in this paper.
%Note that the number of dimensions is not often expected to be large (e.g., in \cite{AI-Feynman} the maximum is $5$).

%\subsection{Architecture}
%The overall architecture of our system is depicted in Figure~\ref{fig:system}.

%\begin{figure}
%\begin{center}
% \centerline{\includegraphics[width=\columnwidth]{fig1.pdf}}
%\includegraphics[width=\columnwidth]{system.png}
%\caption{System Overview {\bf we should remove the part for experimental design}.}
%\label{fig:system}
%\end{center}
%\end{figure}

% Although substantial computational effort was required, we successfully  derived the gravitational constant, whereas the AI-Feynman algorithm lacks a formal mechanism for finding such constants.

\section{Results}

\subsection{Feynman Database for Symbolic Regression}

In our first experiment, we choose a subset of the 100 problems from the Feynman Database for Symbolic Regression (FSReD), created in \cite{AI-Feynman}. We first eliminate the 19 problems which involve operators our system cannot handle because these operators cannot be handled by the underlying MINLP solver we use (namely $\sin, \cos, \arcsin, \tanh$).
We give detailed results on 32 out of the remaining 81 instances with varied properties in Table~\ref{tab1}: some can easily be solved by AI Feynman and Eureqa, and some are hard for these solvers, and some need many data points or limited noise.

\begin{table*}[t] \small
\begin{tabular}{l|l|r|r|r|r|r|r|r}
\hline
Eqn.      & Eqn. & exp 1 & noise     & data  & AIF     & AIF min & AIF max      & Eureqa\\
label     &      &       & exp       & used & time     & data    & noise        & solved \\
\hline
I.6.20a*  & $e^{-\theta^2/2}/\sqrt{2 \pi }$               & 700 &    & 10 & 16             & 10     & $10^{-2}$   & no \\
I.9.18    & $\frac{Gm_1m_2}{(x_2-x_1)^2+(y_2-y_1)^2+(z_2-z_1)^2}$  & f &      & 10  &  5975   &  $10^6$ &  $10^{-5}$  &  no \\
I.10.7   & $\frac{m_0}{\sqrt{1-v^2/c^2}}$             & 11 &      & 10  &  14     &  10     &  $10^{-4}$  &  no \\
I.12.2    & $\frac{q_1q_2}{4\pi \varepsilon r^2}$     & 1 & yes    & 10  &  17     &  10     &  $10^{-2}$  &  yes \\
I.12.4    & $\frac{q_1}{4\pi \varepsilon r^2}$        & 1  &       & 10  &  12     &   10      & $10^{-2}$ & yes \\
I.13.4    & $\frac{1}{2}m(v^2+u^2+w^2)$               & 6 &       & 10  &  22     &  10     &  $10^{-4}$  &  yes\\
I.13.12   & $Gm_1m_2(\frac{1}{r_2}-\frac{1}{r_1})$    & 13 &        & 10  &  20     &  10     &  $10^{-4}$  &  yes \\
I.15.3t   & $\frac{t-ux/c^2}{\sqrt{1-u^2/c^2}}$       & f  &      & 10  &  20     &  $10^2$    &  $10^{-4}$  &  no \\
I.15.10  & $\frac{m_0v}{\sqrt{1-v^2/c^2}}$            & 2  &      & 10  &  13     &  10     &  $10^{-4}$  &  no \\
I.16.6    & $\frac{u+v}{1+uv/c^2}$                   &  10 & {\bf yes}  & 10  &  18     &  10     &  $10^{-3}$  &  no \\
I.18.4   & $\frac{m_1r_1+m_2r_2}{m_1+m_2}$            & 8  &      & 10  &  17     &  10     &  $10^{-2}$  &  yes \\
I.24.6    & $\frac{1}{4}m(\omega^2 + \omega_0^2)x^2$  & 3  &      & 10  &  22     &  10     &  $10^{-4}$  &  yes \\
I.25.13   & $q/C$                                    & 1  &      & 10  &  10     &  10     &  $10^{-2}$  &  yes \\
I.27.6    & $\frac{1}{\frac{1}{d_1}+\frac{n}{d_2}}$  &  5  & yes      & 10  &  14     &  10     &  $10^{-2}$  &  yes \\
I.32.5   & $q^2a^2/(6\pi \varepsilon c^3)$     & f  &      & 10  &  13     &  10     &  $10^{-2}$  &  yes \\
I.34.10   & $\frac{\omega_0}{1-v/c}$                 & 4  &      & 10  &  13     &  10     &  $10^{-3}$  &  no \\
I.34.14   & $\frac{1+v/c}{1-v^2/c^2}\omega_0$        & 7  &      & 10  &  14     &  10     &  $10^{-3}$  &  no \\
I.39.11   & $\frac{1}{\gamma -1}p_FV$                 & 3  &      & 10  &  13     &  10     &  $10^{-3}$  &  yes \\
I.43.43   & $\frac{k_bv}{(\gamma -1)A}$               & 5 &       & 10  &  16    &  $10$ &  $10^{-3}$  &  yes \\
I.48.20   & $\frac{mc^2}{\sqrt{1-v^2/c^2}}$          & 8 &       & {\bf 10}  &  108    &  $10^2$ &  $10^{-5}$  &  no \\
II.2.42   & $k(T_2-T_1)A/d$                  & 13 & {\bf yes} & 10  &  54     &  10     &  $10^{-3}$  &  yes\\
II.11.3   & $\frac{qE_f}{m(\omega_0^2-\omega^2)}$    &  8 & {\bf yes}  & 10  &  25     &  10     &  $10^{-3}$  &  yes\\
II.11.20    & $\frac{n_{\rho}p^2_dE_f}{3k_bT}$        & 9 &            & 10    & 18        &   10      & $10^{-3}$ & yes \\
II.11.27  & $\frac{n\alpha}{1-n\alpha/3}\epsilon E_f$ & 36 & {\bf yes}  & {\bf 10}  &  337    &  $10^2$ &  $10^{-3}$  &  no\\
II.11.28  & $1+\frac{n\alpha}{1-n\alpha/3}$          & f & {\bf yes}   & 10  &  1708   &  $10^2$ &  $10^{-4}$  &  no\\
II.21.32    & $\frac{q}{4\pi \varepsilon r(1-v/c)} $ & 13  &       &  10   &  21             &   10      & $10^{-3}$ & yes \\
II.24.17  & $\sqrt{\frac{\omega^2}{c^2}-\frac{\pi^2}{d^2}}$ &  {\bf 2} & {\bf yes}    & 10  &  62     &  10     &  $10^{-5}$  &  no\\
II.34.11    & $g\_qB/(2m)$                          & 1  &       &  10   &  16             &   10      & $10^{-4}$ & yes \\
II.36.38    & $\frac{\mu_m B}{k_bT} + \frac{\mu_m\alpha M}{\varepsilon c^2k_bT}$  & f  &       &   10  &  77        &   10      & $10^{-2}$ & yes \\
II.37.1    & $\mu_M(1+\chi)B$                        & 1  &       &  10   & 15              &   10      & $10^{-3}$ & yes \\
II.38.3   & $\frac{YAx}{d}$                          & {\bf 1}  & {\bf yes}       & 10  &  47     &  10     &  $10^{-3}$  &  yes\\
III.10.19 & $\mu_M \sqrt{B_x^2+B_y^2+B_z^2}$           &  {\bf 6}  & {\bf yes}       & {\bf 10}  &  410    &  $10^2$ &  $10^{-3}$  &  yes\\
\hline
\end{tabular}
\caption{Results on a subset of problems from the Feynman Database for Symbolic Regression}\label{tab1}
\end{table*}

We set the depth limit $d$ to 3, and the number of constants $k$ to 1. We terminate when we find an expression with objective value (squared error) less than $10^{-4}$ (error tolerance).
The operators we use are described in Section~\ref{sec-desc}. We set $\Omega$ to 100, so all constants are in the range $[-100,100]$.
We set $\delta = 2$ for computational efficiency. This means that each power in an L-monomial lies in the range $[-2,2]$.
Under this setting, the pruning rules -- R1, R2,R3 -- remove potentially non-redundant gentrees,
and our algorithm does not explore all possible symbolic expressions that are representatable as a gentree of depth up to 3. Though L-monomials are closed under multiplication, L-monomials with bounded powers are not.
For example, if there is a solution of the form $x_1^3x_2^2$, we may not find it.

We constrain the search in three more ways.
We bound the sum of the variable powers in an L-monomial by an input number $\tau = 6$.
Thus, for each term of the type (\ref{lmonomial}), we add the constraint $\sum_{i=1}^n|a_i| \leq \tau$, which is representable by a linear constraint using $n$ auxiliary variables $a_i': \sum_{i=1}^n a_i' \leq \tau$ and $a_i' \geq 0, -a_i \leq a_i'$ and $a_i \leq a_i'$ for $i=1, \ldots, n$. 
Secondly, we add another pruning rule. We remove any gentree which contains a subexpression of the form $\sqrt{L}$ where $L$ is an L-monomial (we allow $\sqrt{L_1+L_2}$, for example).
Finally, we drop the operator $-$ from our set. However, as we allow positive and negative constants, we can simulate a $-$ operator.
But because we only allow at most one constant, this means that we effectively allow a single $-$ operator in our generated solutions.
All constants are allowed to have units.

We give our results in Table~\ref{tab1}. The first column contains the equation label from FSReD, the second shows the actual function. The third column gives the running time (in seconds) of our code on a machine with 30 cores, and with $\tau = 6$. A 'f' in this column implies our code did not produce the correct answer within the time limit of 600 seconds. The fourth column gives the outcome of our experiment with noise. The ``data used'' column contains the number of data points we use, which is (the first) 10. Our method does not scale well with the number of data points: equation (\ref{minlp1}) becomes more complex, and the resulting MINLPs harder. We can solve for 20-30 points for most instances, but the running time increasing significantly for some problems. In columns 6-8 we give the time taken by AI Feynman, the min number of points it needs to get the correct answer, and the maximum noise level under which it obtains the correct answer, all taken from the associated paper, as is the Eureqa solution status in the last column.

We fail on 5 instances, but on the remaining instances, our results are mixed in comparison to AI Feynman. We are often slower, but sometimes faster (II.24.17, II.38.3, III.10.19), even if we scale our times by 30 (to account for the parallelism used by our code). We note that AI Feynman searches over more operators, and thus a potentially larger search space. For three instances -- I.48.20, II.11.27, III.10.19 (with the number of data points in bold) -- we obtain the correct answer with one-tenth the number of data points needed by AI Feynman. %In experiment 2, we obtain a speedup owing to more cores, and fact that $\tau = 5$: some equations are expressible with smaller gentrees when $\tau$ is higher. For some problems, we obtain a very significant speedup (I.15.10, I.48.20, II.11.3, II.11.27), but we also fail on I.24.6. We give the time for experiment 1 (resp. experiment 2) in bold if the product of 7 (resp. 30) and this number is less than the AI Feynman running time. 

To solve I.6.20a, indicated by a '*', we first spend the time limit of 600 seconds using the default set of operators, and then restart with $\sqrt{}$ replaced by $\exp()$, and then take another 100 seconds. This is why we report a running time of 700. While using $\exp()$, we also decrease the error tolerance to $10^{-8}$. (We note that AI Feynman also cycles through different sets of operators).
Out of the remaining 49 problems, we are able to solve 36 instances, and fail on 13. Changing some of our parameters will allow additional problems to be solved at the cost of making the code slower. For example, increasing the number of constants $k$ to 2 allows our code to solve I.15.3t. 
Finally, BARON can handle the $\log()$ operator, but we have not experimented with it.

In a second experiment, we set the relative noise level to $10^{-2}$ (as described in the AI Feynman paper), and report a ``yes'' if one of the gentrees returns the expected formula as a solution and has low error. When the ``yes'' is in bold, AI Feynman needs a lower noise level. However, as we generate arbitrary constants, for some problems we find solutions that have lower squared error than the true formula (on the 10 data points we chose), and one should apply AIC or BIC to choose from alternate formulae.

A major difference between our code and AI Feynman is that our code finds arbitrary constants, whereas AI Feynman finds combinations of fixed symbols (such as small integers and $\pi$ and other known physical constants). For example, in I.12.2, we find $1/(4\pi)$ as $0.07957747$. For problem I.13.12,  AI Feynman uses the known value of Newton's Gravitational constant and cannot discover it, whereas our code can.

\section{Conclusions}

A few MINLP based SR methods have been proposed earlier, but these are not very scalable. These methods obtain an ``optimal'' solution and  a certificate of optimality. We proposed an MINLP based approach that enumerates expression trees with fixed operators, and uses MINLP to solve for the form of the leaves. Our approach is competitive with state-of-the-art SR packages on some difficult problems. It incorporates dimensional analysis (a first for an MINLP based implementation) but can also find real-valued constants (as can gplearn). A number of drawbacks -- such as limitations on number of data points, and high variability with respect to parameters -- of our method suggest future directions of research. Even though our overall method is a heuristic, the MINLP solves for individual gentrees return certificates of optimality. One can speedup the method by developing heuristics to find good solutions for a gentree and terminating before proving optimality. 

Our code combines a number of useful features: a) it infers results from limited data, b) discovers complex constants, and c) deals with some noise as in real data, and d) is reasonably fast. AI Feynman cannot handle arbitrary constants or much noise, and Eureqa cannot handle explicit equations (such as dimensional equations, though dimensional equations can be handled implicitly by an application of the the Buckingham-Pi and the implied data preprocessing).\\

\noindent {\bf Acknowledgements} Tyler Josephson was primarily supported by the U.S. Department of Energy (DOE), Office of Basic Energy Sciences, Division of Chemical Sciences, Geosciences and Biosciences under Award DE-FG02-17ER16362. Tyler Josephson and Lior Horesh gratefully acknowledge the support of the Institute for Mathematics and its Applications (IMA), where a part of this work was initiated.

\bibliographystyle{icml2020}
%\bibliography{refs}

\end{document}